\documentclass{article}

\PassOptionsToPackage{numbers,sort&compress}{natbib}
\usepackage[dblblindworkshop,final]{neurips_2026}
\workshoptitle{AXIOM Foundations of Efficient Deep Learning}
\usepackage[utf8]{inputenc}
\usepackage[T1]{fontenc}
\usepackage{hyperref}
\usepackage{url}
\usepackage{booktabs}
\usepackage{amsfonts}
\usepackage{nicefrac}
\usepackage{microtype}
\usepackage{xcolor}
\usepackage{amsmath,amssymb,amsthm}
\usepackage{graphicx}
\usepackage{enumitem}

\newtheorem{theorem}{Theorem}

\newtheorem{corollary}{Corollary}
\newtheorem{proposition}{Proposition}
\theoremstyle{remark}

\newcommand{\NC}{\mathrm{NC}}
\newcommand{\Tr}{\operatorname{Tr}}
\newcommand{\SNR}{\mathrm{SNR}}

\title{A Closed Form Depth-Width Law for Neural Collapse in Deep Linear Networks}

\author{
Pankhil Gawade$^{*,1,6,7}$ \quad
Kirsten Fischer$^{*,1,6}$ \quad
Guy Wolf$^{2,3}$ \quad
Ewa Szczurek$^{1,4,5,6}$
\\[0.8em]
{\small\normalfont
\begin{tabular}{@{}c@{}}
$^1$Institute of AI for Health, Helmholtz Munich, Germany \quad
$^2$Université de Montréal\\
$^3$Mila -- Quebec AI Institute \quad
$^4$Institute of Informatics, University of Warsaw, Warsaw, Poland\\
$^5$Faculty of Mathematics, Informatics and Statistics, LMU Munich, Munich, Germany\\
$^6$Munich Center for Machine Learning (MCML)\\
$^7$School of Computation, Information and Technology, Technical University of Munich\\[0.5em]
\texttt{\{gawade.pankhil, kirsten.fischer, ewa.szczurek\}@helmholtz-munich.de}
\end{tabular}}
}

\begin{document}

\maketitle
\begingroup
\renewcommand{\thefootnote}{*}
\footnotetext{Equal contribution.}
\endgroup

\begin{abstract}
How the depth and width of neural networks shape learned representations is difficult to predict from the data and architecture. We use neural collapse (NC) as a probe of the representation geometry and express its first three properties, within-class variability collapse (NC1), simplex class means (NC2) and classifier alignment (NC3), through the learned feature kernel, extending a known identity for NC1 to NC2 and NC3. For deep linear networks with an exchangeable input kernel, the kernel equations yield a geometric law, exact within this theory, for the layerwise decrease of NC1. At large width $N$ the accumulated decrease becomes a function of the depth-to-width ratio $L/N$, with its rate determined self-consistently by the data and architecture, yielding a scaling law for representation geometry.
\end{abstract}

\section{Introduction}
\label{sec:intro}

Neural collapse (NC) refers to four properties that the last-layer features and the linear classifier of deep classification networks approach in the terminal phase of training, after the training error has vanished: the within-class variability of the features collapses (NC1), the centered class means form a simplex equiangular tight frame (NC2), the classifier vectors align with these class means (NC3), and the classifier reduces to a nearest-class-mean decision (NC4) \citep{papyan2020prevalence}. The ratio of within- to between-class variability also decreases progressively across the layers of trained networks \citep{he2023lawdataseparationdeep,Rangamani23_intermediatenc}. How depth, width and data shape this behavior remains an open question, since most analyses of NC remove these dependencies by treating the layer features as free variables \citep{zhu2021geometric,sukenik2023deep,sukenik2024neural,Hong24_ufm_imbalanced_data}, or constrain the features by keeping them close to their initial values \citep{tirer2023perturbation} or by structural assumptions on the inputs and the weights \citep{Wang25_linearscalinglaw}. With $L$ hidden layers of width $N$, the depth-to-width ratio $L/N$ controls the fluctuations of the neural tangent kernel at initialization \citep{hanin2020finite}, and depth, width and the number $P$ of training samples enter the posterior of Bayesian deep linear networks through $LP/N$ \citep{hanin2023bayesian}. 

Here, we show that $L/N$ also controls the layerwise decrease of NC1 in a solvable setting. To this end, we predict NC across layers from a theory describing the feature kernels in trained networks. Given the labels, NC1--NC3 of a layer depend on its features only through the feature kernel, the Gram matrix of the features over the training samples, if NC3 uses the ridge readout of the layer. For Bayesian networks of finite width, the maximum a posteriori (MAP) feature kernels of all layers follow from self-consistency equations in the data, weight prior and architecture \citep{fischer24critical}. We show that for deep linear networks with an exchangeable input kernel, i.e., invariant under permutations of samples within classes and of the classes, these equations reduce exactly to scalar equations for the within- and between-class kernel eigenvalues on solutions with the same symmetry. Our
contributions are:
\begin{itemize}[leftmargin=1.2em,itemsep=1pt,topsep=2pt,parsep=0pt]
\item kernel forms of NC2 and NC3 (Proposition~\ref{thm:kernel}), complementing that of NC1 \citep{kothapalli2025can}, which we evaluate on the predicted MAP kernels of \citet{fischer24critical} for a nonlinear network (Fig.~\ref{fig:nc123});
\item a geometric law for the layerwise decrease of NC1 in deep linear networks with an exchangeable input kernel (Theorem~\ref{thm:geometric}), exact within this theory, which shows that the accumulated compression is at leading order controlled by the depth-to-width ratio $L/N$, with a rate determined self-consistently by the data, weight prior and architecture (Corollary~\ref{cor:scaling}).
\end{itemize}

\section{Neural collapse observables from feature kernels}
\label{sec:kernel}

Consider a balanced $C$-class problem with $n$ samples per class, $P=Cn$ total training samples and one-hot targets $Y\in\{0,1\}^{P\times C}$. At a fixed layer, the feature vectors $z_\alpha$ of the training samples form the rows of the feature matrix $Z$, with feature kernel $K=ZZ^\top$, class means $\mu_c$, global mean $\mu_G$, and within- and between-class covariances $\Sigma_W$ and $\Sigma_B$. We quantify NC1--NC3 by
\begin{equation}
\NC_1=\frac{\Tr(\Sigma_W)}{\Tr(\Sigma_B)},
\qquad
\NC_2=\big\|\widehat G-\tfrac{C}{C-1}H_C\big\|_F,
\qquad
\NC_3=1-\frac1C\sum_{c=1}^{C}\cos\big(\bar w_c,\mu_c-\mu_G\big).
\label{eq:nc-def}
\end{equation}
For $\NC_1$, we use the trace ratio of \citet{tirer2023perturbation} instead of the measure $\frac1C\Tr(\Sigma_W\Sigma_B^\dagger)$ of
\citet{papyan2020prevalence}.   $\NC_2$ measures the deviation of the cosine Gram matrix $\widehat G_{cc'}=\cos(\mu_c-\mu_G,\mu_{c'}-\mu_G)$ of the centered class means from that of a unit-norm simplex equiangular tight frame $\frac{C}{C-1}H_C$ with $H_C=I_C-\frac1C\mathbf 1_C\mathbf 1_C^\top$.  For $\NC_3$, the trained classifier is replaced by the ridge readout $W=\arg\min_W\|Y^\top-WZ^\top\|_F^2+\kappa\|W\|_F^2$ with ridge parameter $\kappa>0$, whose centered rows are $\bar w_c$.

With $H=I_P-\frac1P\mathbf 1_P\mathbf 1_P^\top$ and $Y_c=HY$, the orthogonal projectors onto the centered class-indicator subspace and onto within-class
fluctuations are
\begin{equation}
P_B = Y_c(Y_c^\top Y_c)^\dagger Y_c^\top,
\qquad
P_W = H-P_B .
\label{eq:projectors}
\end{equation}
The rows of $\widetilde SZ$, with $\widetilde S=H_CY^\top/n$, are the centered
class means, with Gram matrix $G=\widetilde SK\widetilde S^\top$ and
$D=\operatorname{Diag}(G_{11},\ldots,G_{CC})$. We further write
$U=Y^\top(K+\kappa I_P)^{-1}$ and $\widetilde U=H_CU$.

\begin{proposition}[Kernel forms of the neural-collapse observables]
\label{thm:kernel}
For balanced classes and nonvanishing denominators, NC1--NC3 are functions of the feature kernel $K$, the targets $Y$ and, for NC3, the ridge parameter $\kappa$ alone,
\begin{align}
\NC_1 &= \frac{\Tr(P_WK)}{\Tr(P_BK)}, \qquad
\NC_2 = \big\|D^{-1/2}GD^{-1/2}-\tfrac{C}{C-1}H_C\big\|_F ,
\label{eq:nc12}
\\
\NC_3 &= 1-\frac{1}{C}\sum_{c=1}^{C} \frac{\widetilde U_{c:}K\widetilde S_{c:}^{\top}} {\sqrt{\big(\widetilde U_{c:}K\widetilde U_{c:}^{\top}\big) \big(\widetilde S_{c:}K\widetilde S_{c:}^{\top}\big)}},
\label{eq:nc3}
\end{align}
where $A_{c:}$ is row $c$ of $A$. The kernel form of $\NC_1$ is due to \citet[Thm.~4.1]{kothapalli2025can}.
\end{proposition}

\begin{figure}[t]
    \centering
    \includegraphics[width=0.93\linewidth]
    {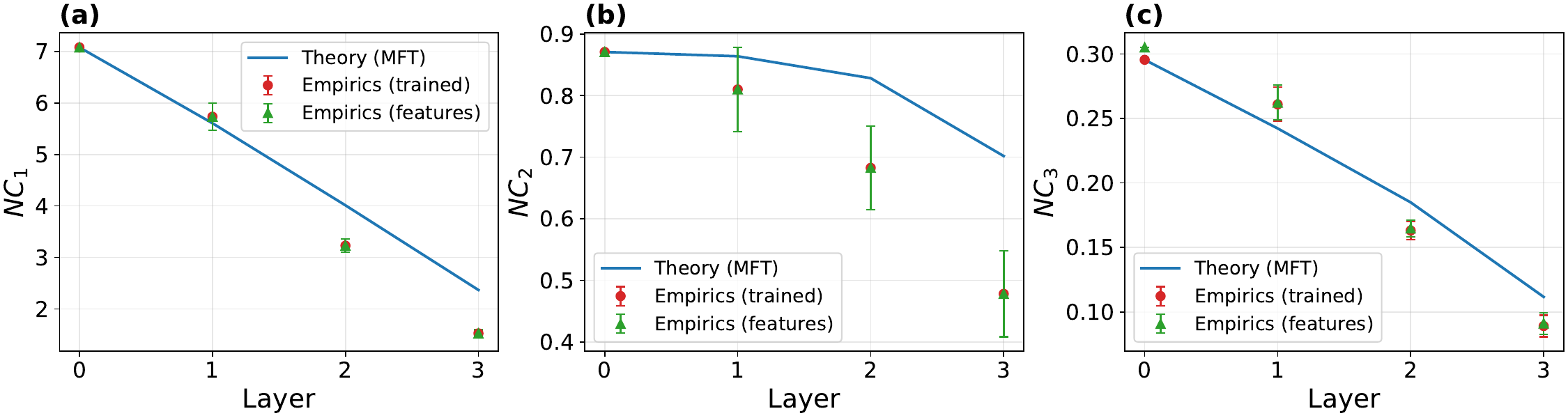}
    \caption{The kernels capture the layerwise decrease of NC1--NC3 in nonlinear networks. NC1--NC3 of an erf network ($N=256$, $L=3$) on four-class XOR from the MAP kernels (lines), the feature kernels $K^{(l)}$ (circles), and raw features $z_{\alpha}^{(l)}$ (triangles) of SGLD-trained networks; error bars over initialization seeds, layer $0$ is the input. Experimental details are in App.~\ref{app:experiments}.}
    \label{fig:nc123}
\end{figure}

Fig.~\ref{fig:nc123} evaluates Prop.~\ref{thm:kernel} on the kernels of a nonlinear network, which qualitatively capture the layerwise decrease of NC1--NC3 in networks trained with stochastic gradient Langevin dynamics (SGLD)~\cite{welling11,naveh21}. NC4, defined on held-out data samples, additionally requires the kernel between test and training inputs (see App.~\ref{app:kernelforms}). These identities hold for arbitrary representations; predicting their layerwise evolution requires a theory of the learned kernels.

\section{Geometric law for deep linear networks with an exchangeable input kernel}
\label{sec:linear}

We consider fully connected networks with $L$ hidden layers of width $N$, $h^{(0)}_\alpha=W^{(0)}x_\alpha+b^{(0)}$ and $h^{(l)}_\alpha=W^{(l)}\phi(h^{(l-1)}_\alpha)+b^{(l)}$ for $l=1,\dots,L$, with activation function $\phi$, inputs $x_\alpha\in\mathbb{R}^{d_0}$ and outputs $h^{(L)}_\alpha\in\mathbb{R}^{C}$. Following \citet{fischer24critical}, the weights have independent Gaussian priors of variance $g_0/d_0$ in the first layer and $g_l/N$ in layer $l$, the biases of variance $g_b$, and the targets have Gaussian observation noise of variance $\kappa$, which we also use as the ridge parameter of NC3. We evaluate Prop.~\ref{thm:kernel} on the kernels $K^{(0)}_{\alpha\beta}=(g_0/d_0)\,x_\alpha^\top x_\beta+g_b$ of the inputs and $K^{(l)}_{\alpha\beta}=(g_l/N)\,z^{(l)\top}_\alpha z^{(l)}_\beta+g_b$ of the features $z^{(l)}_\alpha=\phi(h^{(l-1)}_\alpha)$ of the $l$th hidden layer, giving $\NC_i^{(l)}$; the term $g_b$ adds a constant feature, which leaves NC1 and NC2 unchanged and acts as a readout bias in NC3.

\citet{fischer24critical} derive self-consistency equations for the MAP kernels from a large-deviation principle in $N$ (see App.~\ref{app:map-equations}): forward equations for $K^{(l)}$ and backward equations for conjugate kernels $\widetilde K^{(l)}$, starting from a terminal conjugate kernel. We extend the latter to $N_{\mathrm{out}}=C$ outputs (see App.~\ref{app:multi}), $\widetilde K^{(L)}=\frac12S_L^{-1}(YY^\top-N_{\mathrm{out}} S_L)S_L^{-1}$ with $S_L=K^{(L)}+\kappa I_P$; it measures the mismatch between the target covariance per output, $YY^\top/N_{\mathrm{out}}$, and the output covariance $S_L$. For deep linear networks, $\phi(h)=h$, the equations take the closed form
\begin{equation}
K^{(l)}=g_b\mathbf{1}_P\mathbf{1}_P^\top+g_lK^{(l-1)}T^{(l)},
\qquad
\widetilde K^{(l-1)}=g_lT^{(l)}\widetilde K^{(l)},
\label{eq:exact_d-w}
\end{equation}
with $T^{(l)}=(I_P-\frac{2g_l}{N}\widetilde K^{(l)}K^{(l-1)})^{-1}$ to be solved jointly, as $\widetilde K^{(L)}$ depends on $K^{(L)}$.

\paragraph{Exchangeable input kernel.}
To obtain an analytic solution, we assume that the input kernel is exchangeable, i.e., invariant under simultaneous permutations of its rows and columns that permute the samples within a class or permute the classes, a natural idealization since under NC1 and NC2 the centered kernel $HKH$ is exchangeable. Exchangeable kernels form the commutative algebra $\mathcal A$ of matrices $K= \lambda_EE+\lambda_BP_B+\lambda_WP_W$, with $E=\mathbf 1\mathbf 1^\top/P$ and eigenvalues of multiplicities $1$, $C-1$ and $P-C$. As $\mathbf 1\mathbf 1^\top=PE$ and $YY^\top=n(E+P_B)$ lie in $\mathcal A$, the MAP equations restricted to $\mathcal A$ reduce to scalar equations for the eigenvalues $\lambda_X^{(l)}$ of $K^{(l)}$ and $\widetilde\lambda_X^{(l)}$ of $\widetilde K^{(l)}$, $X\in\{E,B,W\}$ (see App.~\ref{app:exchangeable-algebra}). On such solutions, $\NC_2^{(l)}=\NC_3^{(l)}=0$ and $\NC_1^{(l)}=(P-C)\lambda_W^{(l)}/((C-1)\lambda_B^{(l)})$.

\begin{theorem}[Channel conservation and geometric depth law]
\label{thm:geometric}
Let $\phi(h)=h$, $K^{(0)}\in\mathcal A$ with $\lambda_B^{(0)}>0$ and $\kappa>0$, and consider a solution of the MAP equations in $\mathcal A$. For $X\in\{W,B\}$, $q_X=\lambda_X^{(l)}\widetilde\lambda_X^{(l)}$ is the same for all $l=1,\dots,L$. Hence the eigenvalue $t_X=1+2q_X/N$ of $T^{(l)}$ in channel $X$ is layer-independent and, for $l=0,\dots,L$,
\begin{equation}
\lambda_X^{(l)} = \lambda_X^{(0)}
\textstyle\big(\prod_{m=1}^l g_m\big)t_X^{\,l},
\qquad
\displaystyle
\NC_1^{(l)} = \NC_1^{(0)}
\Big(\frac{t_W}{t_B}\Big)^l .
\label{eq:geometric-law}
\end{equation}
\end{theorem}

The forward and backward equations multiply $\lambda_X$ and $\widetilde\lambda_X$ by the same factor $g_lt_X$ in opposite directions, which conserves their product (see App.~\ref{app:proof-geometric}). In NC1, the common factors $g_l$ cancel, and NC1 decreases geometrically for $t_B>t_W$, in line with the target-dependent kernel corrections of order $1/N$ that grow with depth in deep linear networks \citep{li2021statistical}. The input data fixes $\NC_1^{(0)}$, while the feature-learning corrections, which also depend on the data, set the depthwise contraction.

\begin{corollary}[Depth--width law]
\label{cor:scaling}
Under the assumptions of Theorem~\ref{thm:geometric} and to leading order in $\kappa/\lambda_W^{(L)}$ and $\kappa/\lambda_B^{(L)}$, the compression $\NC_1^{(L)}/\NC_1^{(0)}$ depends on depth and width at large $N$ only through $\tau=L/N$: for $N\to\infty$ at fixed $\tau$ and fixed $\mu=n/(\lambda_B^{(0)}\prod_{m=1}^Lg_m)$,
\begin{equation}
\NC_1^{(L)} / \NC_1^{(0)}
\;\to\;
\exp\!\big[-W_0\big(\mu\tau e^{C\tau}\big)\big]
=
1-\mu\tau+O(\tau^2),
\label{eq:LN}
\end{equation}
with $W_0$ the principal branch of the Lambert function. At finite width, $t_W=1-C/N$ and $t_B=1+(r-C)/N$, where $r=n/\lambda_B^{(L)}$ is, for $N>C$, the unique positive root of $\mu=r\,(1+(r-C)/N)^L$.
\end{corollary}

Since $t_B-t_W=r/N>0$, NC1 decreases in every layer, by the factor $t_W/t_B=1-r/N+O(N^{-2})$. Accumulated over $L$ layers, this gives $\NC_1^{(L)}/\NC_1^{(0)}\approx e^{-rL/N}$, where the rate $r$ itself depends on $L$ and $N$ through the terminal kernel. At large width, $r$ depends on them only through $\tau$, which yields Eq.~\eqref{eq:LN}; for $(\mu+C)\tau\ll1$, it reduces to $\mu$, its value without feature learning (see App.~\ref{app:proof-scaling}). Within the theory, Eq.~\eqref{eq:LN} thus predicts the depth-to-width ratio for a target compression. As in \citet{Wang25_linearscalinglaw}, within-class compression is geometric across layers, but its rate follows here from the feature-learning equations instead of assumptions on the inputs and trained weights.

\paragraph{Input geometry.}
For Gaussian data (see App.~\ref{app:input-geometry}), where $x_\alpha=m_c+\sigma\xi_\alpha$ for a sample $\alpha$ of class $c$, with $\xi_\alpha\sim\mathcal N(0,I_{d_0})$ and centers $m_c$ of equal norms and equal pairwise overlaps, the expected input kernel is exchangeable, with
\begin{equation}
\lambda_W^{(0)}=g_0\sigma^2,
\qquad
\lambda_B^{(0)}=g_0\sigma^2\Big(1+\frac{nC}{C-1}\SNR\Big),
\qquad
\NC_1^{(0)}=\frac{n-1}{n\,\SNR+1-1/C},
\label{eq:input-snr}
\end{equation}
where $\SNR=\sum_c\|m_c-\bar m\|^2/(Cd_0\sigma^2)$ with mean center $\bar m$. Then $\NC_1^{(0)}\to1/\SNR$ for $n\to\infty$.

\begin{figure}[t]
    \centering
    \includegraphics[width=\textwidth]{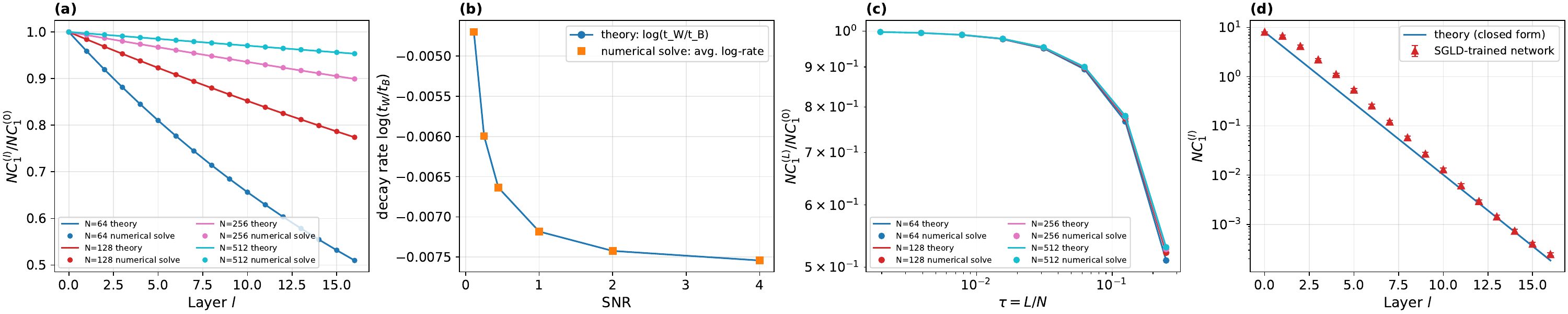}
    \caption{NC1 decreases geometrically with depth and, for the MAP kernels at large width, depends on depth and width only through $L/N$. (a)--(c) Expected input kernel of $C=5$ Gaussian classes with $g_l=1$ for $l<L$; Theorem~\ref{thm:geometric} (lines) and full MAP equations (markers). (a) $\NC_1^{(l)}/\NC_1^{(0)}$ for varying widths and $L=16$. (b) Per-layer compression rate $\log(t_W/t_B)$ versus SNR for $N=256$ and $L=16$. (c) $\NC_1^{(L)}/\NC_1^{(0)}$ versus $\tau=L/N$; the curves nearly coincide. (d) Four sampled Gaussian classes for $N=256$ and $L=16$: SGLD-trained networks (markers) and Theorem~\ref{thm:geometric} for the input kernel projected onto $\mathcal A$ (line), with $g_l=0.5$ for $l<L$. Experimental details in App.~\ref{app:experiments}.}
    \label{fig:linear}
\end{figure}

Figure~\ref{fig:linear} compares these predictions with the numerical solution of the MAP equations and with SGLD-trained networks. On the expected input kernel, theory and numerics agree: NC1 decreases geometrically, more slowly in wider networks (see panel~(a)); in panel~(b), the rate grows with the SNR, here raised by lowering $\sigma$ at fixed centers, which lowers $\lambda_B^{(0)}$; and the terminal compression for $N=64$ to $512$ nearly coincides as a function of $\tau$ (see panel~(c)). SGLD-trained networks follow the per-layer factor of Theorem~\ref{thm:geometric} after a slower decrease in the first layers (see panel~(d)).

\paragraph{Beyond the exchangeable case.}
On four-class XOR, the input kernel is not exchangeable and NC2 and NC3 are nonzero. For an erf network, whose MAP equations we solve to first order in the conjugate kernels (see App.~\ref{app:map-equations}), all three decrease across depth in this theory and in SGLD-trained networks (see Fig.~\ref{fig:nc123}), agreeing best in the first hidden layer, while the trained networks collapse more strongly in deeper layers.

\section{Discussion}
\label{sec:discussion}
In deep linear networks with an exchangeable input kernel, our theory shows how depth, width and data shape the layerwise decrease of NC1: each layer lowers it by the same factor, and at large width, depth and width enter only through $L/N$. Our main limitation is the assumption of input kernel exchangeability, which makes the equations solvable but forces NC2 and NC3 to vanish, so that the law describes NC1 alone and does not hold for general sampled data. Since Proposition~\ref{thm:kernel} applies to any kernel, solving the equations for general input kernels and studying nonlinear networks will be an interesting direction of future research.

\begin{ack}
This work is supported by ERC grant DOG-AMP (No 101125506) [E.S.].
G.W.\ acknowledges support from a Canada CIFAR AI Chair (via Mila), Humboldt Research Fellowship, FRQNT-NSERC grant 2023-NOVA-329125 and
NSERC Discovery grant 03267.

\begin{center}
\includegraphics[height=1.5cm]{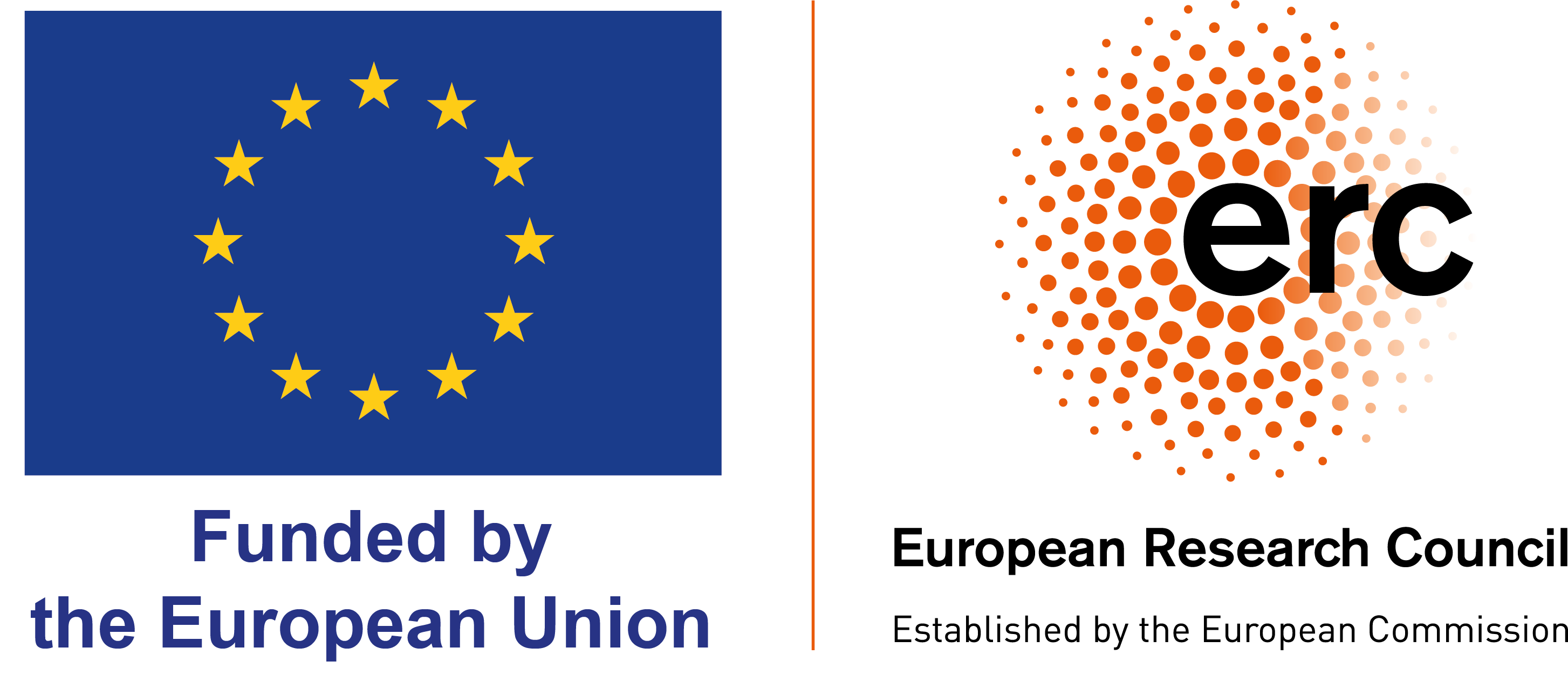}
\end{center}

Funded by the European Union. Views and opinions expressed are however those of the author(s) only and do not necessarily reflect those of the European Union or the European Research Council Executive Agency. Neither the European Union nor the granting authority can be held responsible for them.

\end{ack}

\bibliographystyle{plainnat}
\bibliography{references_v2}

@article{papyan2020prevalence,
author = {Vardan Papyan  and X. Y. Han  and David L. Donoho },
title = {Prevalence of neural collapse during the terminal phase of deep learning training},
journal = {Proceedings of the National Academy of Sciences},
volume = {117},
number = {40},
pages = {24652-24663},
year = {2020},
doi = {10.1073/pnas.2015509117},
URL = {https://www.pnas.org/doi/abs/10.1073/pnas.2015509117},
eprint = {https://www.pnas.org/doi/pdf/10.1073/pnas.2015509117},
}

@article{he2023lawdataseparationdeep,
author = {Hangfeng He  and Weijie J. Su },
title = {A law of data separation in deep learning},
journal = {Proceedings of the National Academy of Sciences},
volume = {120},
number = {36},
pages = {e2221704120},
year = {2023},
doi = {10.1073/pnas.2221704120},
URL = {https://www.pnas.org/doi/abs/10.1073/pnas.2221704120},
eprint = {https://www.pnas.org/doi/pdf/10.1073/pnas.2221704120},
}

@InProceedings{Rangamani23_intermediatenc,
  title = 	 {Feature learning in deep classifiers through Intermediate Neural Collapse},
  author =       {Rangamani, Akshay and Lindegaard, Marius and Galanti, Tomer and Poggio, Tomaso A},
  booktitle = 	 {Proceedings of the 40th International Conference on Machine Learning},
  pages = 	 {28729--28745},
  year = 	 {2023},
  editor = 	 {Krause, Andreas and Brunskill, Emma and Cho, Kyunghyun and Engelhardt, Barbara and Sabato, Sivan and Scarlett, Jonathan},
  volume = 	 {202},
  series = 	 {Proceedings of Machine Learning Research},
  month = 	 {23--29 Jul},
  publisher =    {PMLR},
  url = 	 {https://proceedings.mlr.press/v202/rangamani23a.html},
}

@inproceedings{zhu2021geometric,
 author = {Zhu, Zhihui and Ding, Tianyu and Zhou, Jinxin and Li, Xiao and You, Chong and Sulam, Jeremias and Qu, Qing},
 booktitle = {Advances in Neural Information Processing Systems},
 editor = {M. Ranzato and A. Beygelzimer and Y. Dauphin and P.S. Liang and J. Wortman Vaughan},
 pages = {29820--29834},
 publisher = {Curran Associates, Inc.},
 title = {A Geometric Analysis of Neural Collapse with Unconstrained Features},
 url = {https://proceedings.neurips.cc/paper_files/paper/2021/file/f92586a25bb3145facd64ab20fd554ff-Paper.pdf},
 volume = {34},
 year = {2021}
}

@article{Hong24_ufm_imbalanced_data,
  author  = {Wanli Hong and Shuyang Ling},
  title   = {Neural Collapse for Unconstrained Feature Model under Cross-entropy Loss with Imbalanced Data},
  journal = {Journal of Machine Learning Research},
  year    = {2024},
  volume  = {25},
  number  = {192},
  pages   = {1--48},
  url     = {http://jmlr.org/papers/v25/23-1215.html}
}

@InProceedings{tirer2023perturbation,
  title = 	 {Perturbation Analysis of Neural Collapse},
  author =       {Tirer, Tom and Huang, Haoxiang and Niles-Weed, Jonathan},
  booktitle = 	 {Proceedings of the 40th International Conference on Machine Learning},
  pages = 	 {34301--34329},
  year = 	 {2023},
  editor = 	 {Krause, Andreas and Brunskill, Emma and Cho, Kyunghyun and Engelhardt, Barbara and Sabato, Sivan and Scarlett, Jonathan},
  volume = 	 {202},
  series = 	 {Proceedings of Machine Learning Research},
  month = 	 {23--29 Jul},
  publisher =    {PMLR},
  url = 	 {https://proceedings.mlr.press/v202/tirer23a.html},
}

@article{Wang25_linearscalinglaw,
  author  = {Peng Wang and Xiao Li and Can Yaras and Zhihui Zhu and Laura Balzano and Wei Hu and Qing Qu},
  title   = {Understanding Deep Representation Learning via Layerwise Feature Compression and Discrimination},
  journal = {Journal of Machine Learning Research},
  year    = {2025},
  volume  = {26},
  number  = {220},
  pages   = {1--61},
  url     = {http://jmlr.org/papers/v26/24-0047.html}
}

@article{kothapalli2025can,
  title   = {Can kernel methods explain how the data affects neural collapse?},
  author  = {Kothapalli, Vignesh and Tirer, Tom},
  journal = {Transactions on Machine Learning Research},
  year    = {2025},
  url     = {https://openreview.net/forum?id=MbF1gYfIlY}
}

@InProceedings{fischer24critical,
  title = 	 {Critical feature learning in deep neural networks},
  author =       {Fischer, Kirsten and Lindner, Javed and Dahmen, David and Ringel, Zohar and Kr\"{a}mer, Michael and Helias, Moritz},
  booktitle = 	 {Proceedings of the 41st International Conference on Machine Learning},
  pages = 	 {13660--13690},
  year = 	 {2024},
  editor = 	 {Salakhutdinov, Ruslan and Kolter, Zico and Heller, Katherine and Weller, Adrian and Oliver, Nuria and Scarlett, Jonathan and Berkenkamp, Felix},
  volume = 	 {235},
  series = 	 {Proceedings of Machine Learning Research},
  month = 	 {21--27 Jul},
  publisher =    {PMLR},
  url = 	 {https://proceedings.mlr.press/v235/fischer24a.html},
}

@inproceedings{hanin2020finite,
title={Finite Depth and Width Corrections to the Neural Tangent Kernel},
author={Boris Hanin and Mihai Nica},
booktitle={International Conference on Learning Representations},
year={2020},
url={https://openreview.net/forum?id=SJgndT4KwB}
}

@article{hanin2023bayesian,
author = {Boris Hanin  and Alexander Zlokapa },
title = {Bayesian interpolation with deep linear networks},
journal = {Proceedings of the National Academy of Sciences},
volume = {120},
number = {23},
pages = {e2301345120},
year = {2023},
doi = {10.1073/pnas.2301345120},
URL = {https://www.pnas.org/doi/abs/10.1073/pnas.2301345120},
eprint = {https://www.pnas.org/doi/pdf/10.1073/pnas.2301345120},
}

@article{li2021statistical,
  title = {Statistical Mechanics of Deep Linear Neural Networks: The Backpropagating Kernel Renormalization},
  author = {Li, Qianyi and Sompolinsky, Haim},
  journal = {Phys. Rev. X},
  volume = {11},
  issue = {3},
  pages = {031059},
  numpages = {32},
  year = {2021},
  month = {Sep},
  publisher = {American Physical Society},
  doi = {10.1103/PhysRevX.11.031059},
  url = {https://link.aps.org/doi/10.1103/PhysRevX.11.031059}
}

@inproceedings{sukenik2023deep,
 author = {S\'{u}ken\'{\i}k, Peter and Mondelli, Marco and Lampert, Christoph},
 booktitle = {Advances in Neural Information Processing Systems},
 doi = {10.52202/075280-2306},
 editor = {A. Oh and T. Naumann and A. Globerson and K. Saenko and M. Hardt and S. Levine},
 pages = {52991--53024},
 publisher = {Curran Associates, Inc.},
 title = {Deep Neural Collapse Is Provably Optimal for the Deep Unconstrained Features Model},
 url = {https://proceedings.neurips.cc/paper_files/paper/2023/file/a60c43ba078b723d3d517d28c50ded4c-Paper-Conference.pdf},
 volume = {36},
 year = {2023}
}

@inproceedings{sukenik2024neural,
 author = {S\'{u}ken\'{\i}k, Peter and Lampert, Christoph and Mondelli, Marco},
 booktitle = {Advances in Neural Information Processing Systems},
 doi = {10.52202/079017-4388},
 editor = {A. Globerson and L. Mackey and D. Belgrave and A. Fan and U. Paquet and J. Tomczak and C. Zhang},
 pages = {138250--138288},
 publisher = {Curran Associates, Inc.},
 title = {Neural collapse vs. low-rank bias: Is deep neural collapse really optimal?},
 url = {https://proceedings.neurips.cc/paper_files/paper/2024/file/f9c2ab8d429044e0c35bcece2ff6d123-Paper-Conference.pdf},
 volume = {37},
 year = {2024}
}

@inproceedings{welling11,
author = {Welling, Max and Teh, Yee Whye},
title = {Bayesian learning via stochastic gradient langevin dynamics},
year = {2011},
isbn = {9781450306195},
publisher = {Omnipress},
address = {Madison, WI, USA},
booktitle = {Proceedings of the 28th International Conference on International Conference on Machine Learning},
pages = {681–688},
numpages = {8},
location = {Bellevue, Washington, USA},
series = {ICML'11}
}

@article{naveh21,
  title = {Predicting the outputs of finite deep neural networks trained with noisy gradients},
  author = {Naveh, Gadi and Ben David, Oded and Sompolinsky, Haim and Ringel, Zohar},
  journal = {Phys. Rev. E},
  volume = {104},
  issue = {6},
  pages = {064301},
  numpages = {19},
  year = {2021},
  month = {Dec},
  publisher = {American Physical Society},
  doi = {10.1103/PhysRevE.104.064301},
  url = {https://link.aps.org/doi/10.1103/PhysRevE.104.064301}
}

\newpage
\appendix

\section{Derivation of the kernel neural-collapse observables}
\label{app:kernelforms}

We here prove Proposition~\ref{thm:kernel}, i.e., that NC1--NC3 can be written in terms of the feature kernel $K$, the class labels $y$ and, for NC3, the ridge parameter $\kappa$. We work at an arbitrary fixed layer and make no assumptions on the network architecture, the training dynamics or the exchangeability of the input kernel. The proof consists of three parts. For NC1, label-dependent projectors separate the representation into its centered class-mean and within-class components, whose squared norms are traces of projected kernels. For NC2, the Gram matrix of the centered class means follows by applying the class-averaging and centering operators to $K$. For NC3, the representer form of the ridge readout expresses its alignment with the class means through kernel inner products. The NC1 identity was previously derived by \citet{kothapalli2025can}; we reproduce it in our notation because the same sample-space decomposition is used in the linear analysis of App.~\ref{app:proof-geometric}.

We denote the feature matrix as $Z\in\mathbb{R}^{P\times p}$, whose rows $z_\alpha^\top$ are the representations of the training samples $\alpha=1,\dots,P$, so that $K=ZZ^\top$; the feature dimension is $p=d_0$ at the input and $p=N$ for a hidden layer of the networks in Sec.~\ref{sec:linear}. We consider $C$ balanced classes with $n=P/C$ samples each, labels $y_\alpha\in\{1,\dots,C\}$ and one-hot targets $Y\in\{0,1\}^{P\times C}$. With the centering matrices $H$ and $H_C$ and the centered label matrix $Y_c=HY$ of Sec.~\ref{sec:kernel}, balanced classes imply
\begin{equation}
Y^\top Y=nI_C, \qquad HY=YH_C .
\label{eq:app-balanced}
\end{equation}

\subsection{Proof of the NC1 identity}
\label{app:proof-nc1}

With the class and global feature means
\begin{equation}
\mu_c=\frac1n\sum_{\alpha:y_\alpha=c}z_\alpha,
\qquad
\mu_G=\frac1P\sum_{\alpha=1}^{P}z_\alpha=\frac1C\sum_{c=1}^{C}\mu_c,
\end{equation}
the within- and between-class covariances are
\begin{equation}
\Sigma_W=\frac1P\sum_{c=1}^{C}\sum_{\alpha:y_\alpha=c}(z_\alpha-\mu_c)(z_\alpha-\mu_c)^\top,
\qquad
\Sigma_B=\frac1C\sum_{c=1}^{C}(\mu_c-\mu_G)(\mu_c-\mu_G)^\top .
\label{eq:app-sigma}
\end{equation}
Since $Y_c^\top Y_c=nH_C$ by Eq.~\eqref{eq:app-balanced} and $H_C^\dagger=H_C$, the projector $P_B$ of Eq.~\eqref{eq:projectors} becomes
\begin{equation}
P_B=\frac1nYH_CY^\top=\frac1nYY^\top-\frac1P\mathbf 1_P\mathbf 1_P^\top .
\label{eq:app-pb-expanded}
\end{equation}
We consider a sample $\alpha$ belonging to class $c$, so that row $\alpha$ of $Y$ is the unit vector $e_c^\top\in\mathbb{R}^C$. As the rows of $Y^\top Z$ are the class sums, $(Y^\top Z)_{c:}=\sum_{\beta:y_\beta=c}z_\beta^\top=n\mu_c^\top$, and the rows of $HZ$ are $(HZ)_{\alpha:}=(z_\alpha-\mu_G)^\top$, we obtain
\begin{align}
(P_BZ)_{\alpha:}&=\frac1ne_c^\top H_CY^\top Z=(\mu_c-\mu_G)^\top,
\nonumber\\
(P_WZ)_{\alpha:}&=(HZ)_{\alpha:}-(P_BZ)_{\alpha:}=(z_\alpha-\mu_c)^\top .
\label{eq:app-projector-action}
\end{align}
Thus, $P_BZ$ consists of the centered class means, repeated over the samples of each class, and $P_WZ$ of the within-class residuals. The projector $P_B$ is symmetric and idempotent by construction, and $HP_B=P_BH=P_B$ as $HY_c=Y_c$, so that $P_W$ is symmetric and idempotent as well. Hence $\Tr(P_XK)=\Tr(Z^\top P_XZ)=\|P_XZ\|_F^2$ for $X\in\{B,W\}$, and with Eqs.~\eqref{eq:app-sigma} and \eqref{eq:app-projector-action} and $P=Cn$,
\begin{align}
\Tr(P_BK)&=n\sum_{c=1}^{C}\|\mu_c-\mu_G\|^2=P\,\Tr(\Sigma_B),
\nonumber\\
\Tr(P_WK)&=\sum_{c=1}^{C}\sum_{\alpha:y_\alpha=c}\|z_\alpha-\mu_c\|^2=P\,\Tr(\Sigma_W),
\end{align}
which yields
\begin{equation}
\NC_1=\frac{\Tr(\Sigma_W)}{\Tr(\Sigma_B)}=\frac{\Tr(P_WK)}{\Tr(P_BK)} .
\end{equation}

\subsection{Proof of the NC2 identity}
\label{app:proof-nc2}

With the class-averaging matrix $\Pi=Y/n$, the rows of $\Pi^\top Z$ are the class means $\mu_c^\top$. Centering across classes, $\widetilde S=H_C\Pi^\top=H_CY^\top/n$, gives the centered class means and their Gram matrix,
\begin{equation}
\widetilde SZ=
\begin{bmatrix}
(\mu_1-\mu_G)^\top\\
\vdots\\
(\mu_C-\mu_G)^\top
\end{bmatrix},
\qquad
G=\big(\widetilde SZ\big)\big(\widetilde SZ\big)^\top=\widetilde SK\widetilde S^\top .
\label{eq:app-class-mean-gram}
\end{equation}
If all centered class means are nonzero, $D=\operatorname{Diag}(G_{11},\ldots,G_{CC})$ is invertible, and
$D^{-1/2}GD^{-1/2}$ has the entries
\begin{equation}
\big(D^{-1/2}GD^{-1/2}\big)_{cc'}
=\frac{\langle\mu_c-\mu_G,\mu_{c'}-\mu_G\rangle}
{\|\mu_c-\mu_G\|\,\|\mu_{c'}-\mu_G\|},
\end{equation}
i.e., it equals the cosine Gram matrix $\widehat G$ of the centered class means in Sec.~\ref{sec:kernel}. Since the reference matrix $\frac{C}{C-1}H_C$, the Gram matrix of a unit-norm simplex equiangular tight frame of $C$ vectors, does not depend on the features,
\begin{equation}
\NC_2=\Big\|D^{-1/2}\widetilde SK\widetilde S^\top D^{-1/2}-\frac{C}{C-1}H_C\Big\|_F
\end{equation}
is determined by $K$ and the class labels alone.

\subsection{Proof of the NC3 identity}
\label{app:proof-nc3}

For NC3, we use the bias-free ridge readout $W\in\mathbb{R}^{C\times p}$ of Sec.~\ref{sec:kernel}, which minimizes
\begin{equation}
\big\|Y^\top-WZ^\top\big\|_F^2+\kappa\|W\|_F^2,
\qquad
\kappa>0 .
\end{equation}
The extremum condition $(WZ^\top-Y^\top)Z+\kappa W=0$ gives $W=Y^\top Z(Z^\top Z+\kappa I_p)^{-1}$, and the push-through identity $Z(Z^\top Z+\kappa I_p)^{-1}=(ZZ^\top+\kappa I_P)^{-1}Z$ yields the representer form
\begin{equation}
W=UZ,
\qquad
U=Y^\top\big(K+\kappa I_P\big)^{-1}.
\label{eq:app-representer}
\end{equation}
NC3 compares the class-specific readout directions after removing their common component. The centered readout vectors $\bar w_c^\top$ are the rows of $H_CW=\widetilde UZ$, with $\widetilde U=H_CU$, and the centered class means $\bar\mu_c^\top=(\mu_c-\mu_G)^\top$ are the rows of $\widetilde SZ$. Hence,
\begin{equation}
\langle\bar w_c,\bar\mu_c\rangle=\widetilde U_{c:}K\widetilde S_{c:}^\top,
\qquad
\|\bar w_c\|^2=\widetilde U_{c:}K\widetilde U_{c:}^\top,
\qquad
\|\bar\mu_c\|^2=\widetilde S_{c:}K\widetilde S_{c:}^\top,
\end{equation}
and averaging the cosine misalignment $1-\cos(\bar w_c,\bar\mu_c)$ over classes gives
\begin{equation}
\NC_3=1-\frac1C\sum_{c=1}^{C}
\frac{\widetilde U_{c:}K\widetilde S_{c:}^\top}
{\sqrt{\big(\widetilde U_{c:}K\widetilde U_{c:}^\top\big)\big(\widetilde S_{c:}K\widetilde S_{c:}^\top\big)}},
\qquad
\widetilde U=H_CY^\top\big(K+\kappa I_P\big)^{-1},
\end{equation}
in which every quantity is determined by $K$, $Y$ and $\kappa$. This completes the proof of Proposition~\ref{thm:kernel}.

\subsection{Biases and test samples}
\label{app:kernelforms-remarks}

\paragraph{Biases as a constant feature.}
The kernels $K^{(l)}$ of Sec.~\ref{sec:linear} contain the term $g_b\mathbf 1_P\mathbf 1_P^\top$, i.e., they are the Gram matrices of feature matrices $[Z,\sqrt{g_b}\,\mathbf 1_P]$ with an additional constant column. As $P_B\mathbf 1_P=P_W\mathbf 1_P=\widetilde S\mathbf 1_P=0$, this column changes neither NC1 nor NC2. In NC3, it adds a bias $b\in\mathbb{R}^C$ to the ridge readout, given by $\sqrt{g_b}$ times the readout weights of the constant column, so that the readout is $Wz_\alpha+b$ and the penalty is $\kappa\big(\|W\|_F^2+\|b\|^2/g_b\big)$. The bias is thus regularized with strength $\kappa/g_b$ and also changes $W$ through the joint fit. As the constant column vanishes in the centered class means, the bias does not enter the inner products $\langle\bar w_c,\bar\mu_c\rangle$, but it raises the squared norms of the centered readout vectors from $\|\bar w_c\|^2$ to $\|\bar w_c\|^2+\bar b_c^2/g_b$, with $\bar b=H_Cb$, which enter the denominators in Eq.~\eqref{eq:nc3}.

\paragraph{NC4.}
On a test input $x$ with features $z(x)$, NC4 compares the readout decision $\arg\max_c w_c^\top z(x)$ with the nearest class mean
$\arg\min_c\|z(x)-\mu_c\|$. Both decisions require the kernel between the test input and the training inputs, which is not contained in $K$. We therefore do not give a kernel form of NC4.

\section{MAP kernel equations}
\label{app:map-equations}

\citet{fischer24critical} determine the most probable kernels of a network of large width $N$ in a Bayesian setting, conditioned on its training data. Under the weight prior and given $K^{(l-1)}$, the preactivation vectors $h\in\mathbb{R}^P$ of the $N$ neurons of layer $l-1$ across the training samples are independent and Gaussian with covariance $K^{(l-1)}$, and $K^{(l)}-g_b\mathbf 1_P\mathbf 1_P^\top$ is $g_l$ times the average of $\phi(h)\phi(h)^\top$ over these neurons. Its distribution given $K^{(l-1)}=K$ thus obeys a large-deviation principle in $N$ with the cumulant-generating function
\begin{equation}
\mathcal W_l\big(\widetilde K\,\big|\,K\big)
=N\ln\Big\langle\exp\Big(\frac{g_l}{N}\,\phi(h)^\top\widetilde K\,\phi(h)\Big)\Big\rangle_{h\sim\mathcal N(0,K)},
\label{eq:app-cgf}
\end{equation}
whose argument $\widetilde K\in\mathbb{R}^{P\times P}$ is the conjugate kernel, the variable conjugate to $K^{(l)}$ in the Legendre transform of $\mathcal W_l$ that gives the rate function of $K^{(l)}$. The stationary point of the resulting log posterior of the kernels obeys forward equations $K^{(l)}=g_b\mathbf 1_P\mathbf 1_P^\top+\partial\mathcal W_l(\widetilde K^{(l)}|K^{(l-1)})/\partial\widetilde K^{(l)}$ and backward equations $\widetilde K^{(l-1)}=\partial\mathcal W_l(\widetilde K^{(l)}|K^{(l-1)})/\partial K^{(l-1)}$. They start from the input kernel $K^{(0)}$ and from the terminal conjugate kernel, which, for a single output with targets $Y\in\mathbb{R}^{P\times1}$ and Gaussian observation noise of variance $\kappa$, is
\begin{equation}
\widetilde K^{(L)}=\frac12S_L^{-1}\big(YY^\top-S_L\big)S_L^{-1},
\qquad
S_L=K^{(L)}+\kappa I_P.
\label{eq:app-terminal-scalar}
\end{equation}
As $\widetilde K^{(L)}$ depends on $K^{(L)}$, the equations must be solved jointly. The forward equation evaluates $g_l\langle\phi\phi^\top\rangle$ under the Gaussian measure with covariance $K^{(l-1)}$, tilted by $\exp(g_l\phi^\top\widetilde K^{(l)}\phi/N)$, and reduces to the recursion for the kernels of the neural network Gaussian process (NNGP) for $\widetilde K^{(l)}=0$. For linear networks, the tilted measure is Gaussian, so that the equations close exactly in the kernels; for nonlinear networks, it is not, and we expand the equations in the conjugate kernels.

\paragraph{Linear networks.}
\label{app:linear-map-equations}
For $\phi(h)=h$, Eq.~\eqref{eq:app-cgf} is a Gaussian integral, $\mathcal W_l(\widetilde K|K)=-\frac N2\ln\det(I_P-\frac{2g_l}{N}\widetilde KK)$, and its derivatives give, without approximation and for $l=1,\dots,L$,
\begin{equation}
K^{(l)}=g_b\mathbf 1_P\mathbf 1_P^\top+g_lK^{(l-1)}T^{(l)},
\qquad
\widetilde K^{(l-1)}=g_lT^{(l)}\widetilde K^{(l)},
\label{eq:app-transfer-form}
\end{equation}
with the transfer matrix $T^{(l)}=(I_P-\frac{2g_l}{N}\widetilde K^{(l)}K^{(l-1)})^{-1}$. These are the equations of Sec.~\ref{sec:linear}, Eq.~\eqref{eq:exact_d-w}. The same matrix $T^{(l)}$ acts from the right on the kernel and from the left on the conjugate kernel.

\paragraph{Nonlinear networks.}
For nonlinear $\phi$, the tilted expectations have no closed form. Following \citet[Sec.~3.4]{fischer24critical}, we expand the forward and backward equations to first order in the conjugate kernels, assuming $P/N\ll1$, and have
\begin{align}
K^{(l)}_{\alpha\beta}&=g_b+g_l\langle\phi_\alpha\phi_\beta\rangle
+\frac{g_l^2}{N}\sum_{\gamma,\delta}V^{(l-1)}_{\alpha\beta,\gamma\delta}\,\widetilde K^{(l)}_{\gamma\delta},
\label{eq:app-forward-nonlinear}\\
\widetilde K^{(l-1)}_{\alpha\beta}&=g_l\,\widetilde K^{(l)}_{\alpha\beta}\,\langle\phi'_\alpha\phi'_\beta\rangle
+\delta_{\alpha\beta}\,g_l\sum_{\gamma}\widetilde K^{(l)}_{\alpha\gamma}\,\langle\phi''_\alpha\phi_\gamma\rangle,
\label{eq:app-backward-nonlinear}
\end{align}
with $\phi_\alpha=\phi(h_\alpha)$, all expectations over $h\sim\mathcal N(0,K^{(l-1)})$, and the connected four-point function $V^{(l-1)}_{\alpha\beta,\gamma\delta}=\langle\phi_\alpha\phi_\beta\phi_\gamma\phi_\delta\rangle
-\langle\phi_\alpha\phi_\beta\rangle\langle\phi_\gamma\phi_\delta\rangle$.
Together with the terminal condition for $C$ outputs, Eq.~\eqref{eq:app-multi-terminal}, Eqs.~\eqref{eq:app-forward-nonlinear} and \eqref{eq:app-backward-nonlinear} give the theory curves in Fig.~\ref{fig:nc123} (see App.~\ref{app:experiments} for more details).

\section{Terminal condition for multidimensional network outputs}
\label{app:multi}

The feature-learning theory of \citet{fischer24critical} is formulated for a scalar network output, which covers binary classification. To study neural collapse in the more general $C$-class case with one-hot encoded labels, we here extend the terminal condition to $N_{\mathrm{out}}$-dimensional outputs. This condition expresses the terminal conjugate kernel $\widetilde K^{(L)}$, from which the backward equations start, through $K^{(L)}$ and the targets, and thereby enters all results below; for one-hot targets, it is the origin of the factor $C$ in the transfer factors of Corollary~\ref{cor:scaling}.

The forward equations for the feature kernels and the backward equations for the conjugate kernels are unchanged: all $N_{\mathrm{out}}$ outputs share the same feature kernel $K^{(L)}\in\mathbb{R}^{P\times P}$. The output dimension enters only through the data term $\mathcal{S}_D$ of the action, the log posterior of the kernels \citep[Eq.~(9)]{fischer24critical}.

We denote the network output for training sample $\alpha$ by $f_\alpha\in\mathbb{R}^{N_{\mathrm{out}}}$. The readout weights $W^{(L)}\in\mathbb{R}^{N_{\mathrm{out}}\times N}$ and biases $b^{(L)}\in\mathbb{R}^{N_{\mathrm{out}}}$ have independent rows and entries, respectively,
\[
W^{(L)}_{k\cdot}\sim\mathcal{N}\big(0,\tfrac{g_L}{N}I_N\big), \qquad
b^{(L)}_k\sim\mathcal{N}(0,g_b), \qquad
k=1,\ldots,N_{\mathrm{out}}.
\]
The targets are collected in $Y\in\mathbb{R}^{P\times N_{\mathrm{out}}}$. For one-hot encoded labels $y_\alpha\in\{1,\ldots,C\}$, the targets are $Y_{\alpha k}=\mathbf{1}\{y_\alpha=k\}$ and $N_{\mathrm{out}}=C$. Assuming independent Gaussian observation noise of variance $\kappa$ for each output coordinate, the likelihood is
\begin{equation}
p(Y\mid f) = \prod_{\alpha=1}^{P} \prod_{k=1}^{N_{\mathrm{out}}} \mathcal{N}\big(Y_{\alpha k}\mid f_{\alpha k},\kappa\big).
\label{eq:app-multi-noise}
\end{equation}

Conditioned on the activations of the last hidden layer, the $N_{\mathrm{out}}$ outputs are independent Gaussian processes on the training inputs with the common covariance $K^{(L)}$. Marginalizing the readout weights and biases therefore gives, with $S_L=K^{(L)}+\kappa I_P$ as in Sec.~\ref{sec:linear},
\begin{align}
p\big(Y\mid K^{(L)}\big) 
&= \prod_{k=1}^{N_{\mathrm{out}}} \mathcal{N}\big(Y_{\cdot k}\mid 0,S_L\big), \nonumber\\ \mathcal{S}_D = \ln p\big(Y\mid K^{(L)}\big)
&= -\frac12\Tr\big(S_L^{-1}YY^\top\big) -\frac{N_{\mathrm{out}}}{2}\ln\det S_L +\mathrm{const}.
\label{eq:app-multi-evidence}
\end{align}
Relative to the scalar-output case, the trace term keeps its form, while the log-determinant acquires the factor $N_{\mathrm{out}}$.

Stationarity of the action with respect to $K^{(L)}$, together with the equation of state, gives $\widetilde K^{(L)}=\partial\mathcal{S}_D/\partial K^{(L)}$. Differentiating Eq.~\eqref{eq:app-multi-evidence} with respect to $K^{(L)}$, using $\partial\ln\det S_L/\partial K^{(L)}=S_L^{-1}$ and $\partial\Tr(S_L^{-1}YY^\top)/\partial K^{(L)}=-S_L^{-1}YY^\top S_L^{-1}$, yields
\begin{equation}
\widetilde K^{(L)} = \frac12 S_L^{-1} \big(YY^\top-N_{\mathrm{out}}S_L\big) S_L^{-1}.
\label{eq:app-multi-terminal}
\end{equation}
Thus, relative to the scalar-output terminal condition, only the subtracted term acquires the factor $N_{\mathrm{out}}$; the two outer resolvents are unchanged. Equation~\eqref{eq:app-multi-terminal} reduces to Eq.~\eqref{eq:app-terminal-scalar} at $N_{\mathrm{out}}=1$. From here on, we use one-hot targets, $N_{\mathrm{out}}=C$. Equation~\eqref{eq:app-multi-terminal} is therefore the only modification
required for the multiclass analysis below.

\section{Proof of Theorem~\ref{thm:geometric}}
\label{app:proof-geometric}

We here prove Theorem~\ref{thm:geometric}; Corollary~\ref{cor:scaling} is proven in App.~\ref{app:proof-scaling}. The proof has two steps. First, for an exchangeable input kernel, we show that the linear MAP equations, Eqs.~\eqref{eq:app-transfer-form} and \eqref{eq:app-multi-terminal}, restricted to the algebra $\mathcal A$, decouple into independent scalar equations for the channels $E$, $B$ and $W$ of the global mean, the centered class means and the within-class fluctuations, and that NC2 and NC3 vanish on such solutions (App.~\ref{app:exchangeable-algebra}). Second, the scalar forward and backward equations conserve the product of the kernel and conjugate-kernel eigenvalues in each channel, which makes the transfer factor independent of the layer; the geometric law for NC1 then follows from the ratio of the within- and between-class eigenvalues (App.~\ref{app:conservation}).

\subsection{Exchangeable input kernel}
\label{app:exchangeable-algebra}

Equations~\eqref{eq:app-transfer-form} and \eqref{eq:app-multi-terminal} are matrix equations for $P\times P$ kernels. To solve them in closed form, we now assume that the input kernel is exchangeable, i.e., invariant under simultaneous permutations of its rows and columns that permute the samples within a class or permute the classes. This is a strong assumption on the input geometry, which sampled data satisfy approximately at best. We adopt it because it makes the coupled forward and backward equations analytically solvable, and because the solution isolates the mechanism behind the layerwise decrease of $\NC_1$.

An exchangeable kernel has only three distinct entries: a diagonal value $k_0$, a within-class off-diagonal value $k_w$, and a between-class value $k_b$. Any such kernel can be written as
\begin{equation}
K=(k_0-k_w)\,I_P+(k_w-k_b)\,YY^\top+k_b\,\mathbf{1}_P\mathbf{1}_P^\top .
\label{eq:app-exchangeable-entry-decomposition}
\end{equation}

\paragraph{Projector basis.}
The three entry types are resolved by three mutually orthogonal projectors that sum to the identity,
\begin{equation}
E=\frac{1}{P}\mathbf{1}_P\mathbf{1}_P^{\top},\qquad
P_B=\frac{1}{n}YY^\top-E,\qquad
P_W=I_P-\frac{1}{n}YY^\top ,
\label{eq:app-exchangeable-projectors}
\end{equation}
which extract, respectively, the global-mean direction (rank $1$), the centered class-mean subspace (rank $C-1$), and the within-class fluctuations (rank $P-C$). As $Y_c=HY$ satisfies $Y_c^\top Y_c=nH_C$, $P_B$ and $P_W$ coincide with the projectors of Eq.~\eqref{eq:projectors}. In this basis, Eq.~\eqref{eq:app-exchangeable-entry-decomposition} becomes
\begin{equation}
K=\lambda_EE+\lambda_BP_B+\lambda_WP_W,
\end{equation}
with the eigenvalues
\begin{equation}
\lambda_E=k_0+(n-1)k_w+n(C-1)k_b,
\qquad
\lambda_B=k_0-k_w+n(k_w-k_b),
\qquad
\lambda_W=k_0-k_w .
\label{eq:app-exchangeable-eigenvalues}
\end{equation}
The channel $E$ collects the uniform component of the kernel. It drops out of NC1 and NC2, as $P_B\mathbf 1_P=P_W\mathbf 1_P=\widetilde S\mathbf 1_P=0$, and for kernels in $\mathcal A$, NC3 vanishes for any $\lambda_E$ (see below), so that we track only the channels $B$ and $W$ in what follows.

\paragraph{Invariance of the equations.}
Exchangeable kernels form the commutative algebra $\mathcal A=\operatorname{span}\{E,P_B,P_W\}$, which is closed under the sums, products and inverses that appear in the forward and backward equations. The two remaining ingredients also lie in $\mathcal A$: the bias term is purely global-mean, $g_b\mathbf{1}_P\mathbf{1}_P^\top=g_bP\,E$, and the target kernel occupies the global-mean and between-class channels,
\begin{equation}
YY^\top=n(E+P_B),
\qquad
(y_E,\,y_B,\,y_W)=(n,\,n,\,0).
\label{eq:app-target-projector-decomposition}
\end{equation}
Restricted to $\mathcal A$, the equations therefore close: if all $K^{(l)}$ and $\widetilde K^{(l)}$ lie in $\mathcal A$, so do the transfer matrices $T^{(l)}$, and the matrix equations reduce to scalar equations for the eigenvalues. We consider such solutions; as the equations are only equivariant under the permutations, solutions outside $\mathcal A$ are not excluded. Within $\mathcal A$, $\widetilde K^{(l)}$ and $K^{(l-1)}$ commute, so that $T^{(l)}$ is symmetric and its left and right actions in Eq.~\eqref{eq:app-transfer-form} coincide: a single scalar $t_X^{(l)}$ per channel governs the forward and backward equations, which decouple into three independent scalar channels.

\paragraph{NC2 and NC3.}
On solutions in $\mathcal A$, NC2 and NC3 vanish at every layer with $\lambda_B>0$. With $\widetilde S=H_CY^\top/n$ of Sec.~\ref{sec:kernel} and the projectors of Eq.~\eqref{eq:app-exchangeable-projectors}, $Y^\top Y=nI_C$ and $Y^\top\mathbf 1_P=n\mathbf 1_C$ imply $Y^\top P_W=0$, $H_CY^\top E=0$ and $H_CY^\top P_B=H_CY^\top$, so that $\widetilde SK=\lambda_B\widetilde S$ for every $K\in\mathcal A$. Hence $G=\widetilde SK\widetilde S^\top=(\lambda_B/n)H_C$ and $D=(\lambda_B/n)(1-1/C)\,I_C$, so that $D^{-1/2}GD^{-1/2}=\frac{C}{C-1}H_C$ and $\NC_2=0$. Likewise, $(K+\kappa I_P)^{-1}\in\mathcal A$ has the eigenvalue $(\lambda_B+\kappa)^{-1}$ in channel $B$, so that $\widetilde U=H_CY^\top(K+\kappa I_P)^{-1}=n\widetilde S/(\lambda_B+\kappa)$. Every row $\widetilde U_{c:}$ is thus a positive multiple of $\widetilde S_{c:}$, as $\lambda_B+\kappa>0$, all cosines in Eq.~\eqref{eq:nc3} equal one, and $\NC_3=0$.

\subsection{Conservation law and geometric depth law}
\label{app:conservation}

Writing the eigenvalues of $K^{(l)}$, $\widetilde K^{(l)}$ and $T^{(l)}$ in channel $X\in\{B,W\}$ as $\lambda_X^{(l)}$, $\widetilde\lambda_X^{(l)}$ and $t_X^{(l)}$, Eq.~\eqref{eq:app-transfer-form} reduces per channel to
\begin{equation}
\lambda_X^{(l)}=g_l\,\lambda_X^{(l-1)}\,t_X^{(l)},
\qquad
\widetilde\lambda_X^{(l-1)}=g_l\,\widetilde\lambda_X^{(l)}\,t_X^{(l)},
\qquad
t_X^{(l)}=\Big(1-\frac{2g_l}{N}\,\widetilde\lambda_X^{(l)}\lambda_X^{(l-1)}\Big)^{-1},
\label{eq:app-scalar-channels}
\end{equation}
where the bias contributes only to the channel $E$ and hence drops from both $B$ and $W$.

\begin{proof}[Proof of Theorem~\ref{thm:geometric}]
Inserting the forward recursion of Eq.~\eqref{eq:app-scalar-channels} into $\lambda_X^{(l)}\widetilde\lambda_X^{(l)}$ and the backward recursion into $\lambda_X^{(l-1)}\widetilde\lambda_X^{(l-1)}$ gives the same expression, $g_l\,t_X^{(l)}\lambda_X^{(l-1)}\widetilde\lambda_X^{(l)}$: the forward and backward eigenvalues change by the same factor $g_l t_X^{(l)}$ in opposite directions, so that their product is conserved, $q_X=\lambda_X^{(l)}\widetilde\lambda_X^{(l)}$ for all $l=1,\dots,L$. Substituting $\widetilde\lambda_X^{(l)}\lambda_X^{(l-1)}=q_X/(g_lt_X^{(l)})$ into the definition of $t_X^{(l)}$ gives $t_X^{(l)}=\big(1-2q_X/(Nt_X^{(l)})\big)^{-1}$, which rearranges to
\begin{equation}
t_X^{(l)}=1+\frac{2q_X}{N}\equiv t_X .
\label{eq:app-layer-independent-transfer}
\end{equation}
The transfer factor is thus independent of the layer, and iterating the forward recursion yields, for $l=0,\dots,L$,
\begin{equation}
\lambda_X^{(l)}=\lambda_X^{(0)}\Big(\prod_{m=1}^{l}g_m\Big)t_X^{\,l}.
\label{eq:app-channel-evolution}
\end{equation}
NC1 in Eq.~\eqref{eq:nc12} is the ratio of within- to between-class kernel mass, $\NC_1^{(l)}=\Tr(P_WK^{(l)})/\Tr(P_BK^{(l)})$. Since $K^{(l)}$ is scalar on each projector subspace and the trace of an orthogonal projector equals its rank,
\begin{equation}
\Tr(P_WK^{(l)})=(P-C)\,\lambda_W^{(l)},
\qquad
\Tr(P_BK^{(l)})=(C-1)\,\lambda_B^{(l)},
\label{eq:app-within-kernel-trace}
\end{equation}
so that
\begin{equation}
\NC_1^{(l)}=\frac{P-C}{C-1}\,\frac{\lambda_W^{(l)}}{\lambda_B^{(l)}}.
\label{eq:app-linear-nc1-ratio}
\end{equation}
Inserting Eq.~\eqref{eq:app-channel-evolution}, the factor $\prod_{m=1}^{l}g_m$, which is the same for both channels, cancels, and
\begin{equation}
\NC_1^{(l)}
=\frac{P-C}{C-1}\,\frac{\lambda_W^{(0)}}{\lambda_B^{(0)}}\Big(\frac{t_W}{t_B}\Big)^{l}
=\NC_1^{(0)}\Big(\frac{t_W}{t_B}\Big)^{l},
\label{eq:app-exact-geometric-nc1}
\end{equation}
where the prefactor is $\NC_1^{(0)}$ by Eq.~\eqref{eq:app-linear-nc1-ratio} at $l=0$.
\end{proof}

The decrease of NC1 is therefore controlled entirely by the ratio $t_W/t_B$: the input kernel fixes the prefactor $\NC_1^{(0)}$, while feature learning drives the layerwise decrease through the two transfer factors. Within-class fluctuations are compressed relative to the between-class signal whenever $t_W<t_B$, and the per-layer factor is constant, so that NC1 decreases geometrically with depth. The transfer factors also depend on the input data through the terminal condition, which we analyze in detail in the next section.

\section{Proof of Corollary~\ref{cor:scaling}}
\label{app:proof-scaling}

We next derive the transfer factors at small $\kappa$ and the depth--width law of Corollary~\ref{cor:scaling}. We  denote the terminal eigenvalue in channel $X$ as $\Lambda_X=\lambda_X^{(L)}$. Since the terminal condition, Eq.~\eqref{eq:app-multi-terminal}, is diagonal in the projector basis, it becomes
\begin{equation}
\widetilde\lambda_X^{(L)}=\frac{1}{2}\,\frac{y_X-C(\Lambda_X+\kappa)}{(\Lambda_X+\kappa)^2},
\label{eq:app-terminal-channel}
\end{equation}
where $y_X$ is the eigenvalue of $YY^\top$ in channel $X$ from Eq.~\eqref{eq:app-target-projector-decomposition}.

\subsection{Exact terminal closure}
Before any approximation, we give the exact self-consistency equations for the transfer factors. Combining $t_X=1+2q_X/N$ with $q_X=\Lambda_X\widetilde\lambda_X^{(L)}$ and Eq.~\eqref{eq:app-terminal-channel} expresses the transfer factor as an explicit function of the terminal eigenvalue alone,
\begin{equation}
t_X(\Lambda_X)=1+\frac{1}{N}\,\frac{\Lambda_X\big[y_X-C(\Lambda_X+\kappa)\big]}{(\Lambda_X+\kappa)^2}.
\label{eq:app-transfer-exact}
\end{equation}
The eigenvalue $\Lambda_X$ is in turn the terminal value of Eq.~\eqref{eq:app-channel-evolution}, $\Lambda_X=\Lambda_X^{\mathrm{NNGP}}t_X^{L}$, with the NNGP eigenvalue $\Lambda_X^{\mathrm{NNGP}}=\lambda_X^{(0)}\prod_{m=1}^{L}g_m$, for which $\widetilde K^{(l)}=0$ and $t_X=1$. Equation~\eqref{eq:app-transfer-exact} thus closes into a single scalar fixed-point equation for the terminal eigenvalue,
\begin{equation}
\Lambda_X=\Lambda_X^{\mathrm{NNGP}}\bigg(1+\frac{1}{N}\,\frac{\Lambda_X\big[y_X-C(\Lambda_X+\kappa)\big]}{(\Lambda_X+\kappa)^2}\bigg)^{L}.
\label{eq:app-terminal-fixed-point}
\end{equation}
This closure is exact within the MAP theory, as the geometric profile and the conservation law hold at any finite width. Admissible roots have $t_X>0$, for which the Gaussian integral in Eq.~\eqref{eq:app-cgf} converges; several such roots can occur for $\Lambda_X^{\mathrm{NNGP}}\ll\kappa$. Through $\Lambda_X^{\mathrm{NNGP}}$, the solution depends on the input kernel via $\lambda_X^{(0)}$, on the weight prior via $\prod_mg_m$, and on depth and width explicitly; this dependence carries the input data into the per-layer factor. We now simplify the transfer factors for small $\kappa$.

\subsection{Small-\texorpdfstring{$\kappa$}{kappa} transfer factors}
\label{app:small-kappa-transfer}

\paragraph{Within-class channel.}
The targets have no within-class component, $y_W=0$. Hence,
\begin{equation}
\widetilde\lambda_W^{(L)}=-\frac{C}{2}\,\frac{1}{\Lambda_W+\kappa},
\qquad
q_W=\Lambda_W\widetilde\lambda_W^{(L)}=-\frac{C}{2}\,\frac{\Lambda_W}{\Lambda_W+\kappa}.
\label{eq:app-qw-exact}
\end{equation}
For $\kappa\ll\Lambda_W$, $q_W=-C/2+O(C\kappa/\Lambda_W)$, so that
\begin{equation}
t_W=1-\frac{C}{N}+O\Big(\frac{C\kappa}{N\Lambda_W}\Big).
\label{eq:app-tw-small-kappa}
\end{equation}
Thus, the within-class variance is suppressed by a per-layer factor that deviates from one by $C/N$ at leading order.

\paragraph{Between-class channel.}
The eigenvalue of the one-hot targets in the between-class channel is $y_B=n$. With the ratio of target to terminal between-class variance
\begin{equation}
r=\frac{y_B}{\Lambda_B}=\frac{n}{\Lambda_B},
\label{eq:app-r-definition}
\end{equation}
Eq.~\eqref{eq:app-terminal-channel} gives
\begin{equation}
q_B=\Lambda_B\widetilde\lambda_B^{(L)}=\frac{1}{2}\,\frac{\Lambda_B\big[n-C(\Lambda_B+\kappa)\big]}{(\Lambda_B+\kappa)^2}.
\label{eq:app-qb-exact}
\end{equation}
For $\kappa\ll\Lambda_B$,
\begin{equation}
q_B=\frac{1}{2}(r-C)+O\Big(\frac{(r+C)\kappa}{\Lambda_B}\Big),
\label{eq:app-qb-small-kappa}
\end{equation}
and therefore
\begin{equation}
t_B=1+\frac{r-C}{N}+O\Big(\frac{(r+C)\kappa}{N\Lambda_B}\Big).
\label{eq:app-tb-small-kappa}
\end{equation}

\subsection{Large-width depth--width scaling}
Neglecting the corrections in $\kappa$ and assuming $N>C$, the ratio of the transfer factors is
\begin{equation}
\frac{t_W}{t_B}=\frac{1-C/N}{1+(r-C)/N}=\frac{N-C}{N+r-C}=1-\frac{r}{N+r-C}.
\label{eq:app-transfer-ratio-exact}
\end{equation}
For $r\ll N$, this gives
\begin{equation}
\frac{t_W}{t_B}=1-\frac{r}{N}+O\Big(\frac{r(r+C)}{N^2}\Big),
\qquad
\ln\Big(\frac{t_W}{t_B}\Big)=-\frac{r}{N}+O\Big(\frac{r(r+C)}{N^2}\Big),
\label{eq:app-transfer-ratio-expansion}
\end{equation}
and substitution into Theorem~\ref{thm:geometric} yields
\begin{equation}
\NC_1^{(L)}=\NC_1^{(0)}\exp\Big[-\frac{rL}{N}+O\Big(\frac{Lr(r+C)}{N^2}\Big)\Big].
\label{eq:app-depth-width-expansion}
\end{equation}
The explicit dependence on $C$ cancels from the leading rate $r/N$; the number of classes enters only through $r$, which we determine below.

\subsection{Self-consistent determination of the rate}
\label{app:rate}

We next determine $r$ at small $\kappa$. With the NNGP eigenvalue of the between-class channel $\Lambda_B^{\mathrm{NNGP}}=\lambda_B^{(0)}\prod_{m=1}^{L}g_m$, we define the target drive
\begin{equation}
\mu=\frac{n}{\Lambda_B^{\mathrm{NNGP}}},
\label{eq:app-between-drive}
\end{equation}
the value of $r$ for the NNGP kernels. With $\Lambda_B=\Lambda_B^{\mathrm{NNGP}}t_B^L$ and $r=n/\Lambda_B$, Eq.~\eqref{eq:app-terminal-fixed-point} for $X=B$ is equivalent to $\mu=r\,t_B^L$, which at small $\kappa$ with Eq.~\eqref{eq:app-tb-small-kappa} becomes
\begin{equation}
\mu=r\Big(1+\frac{r-C}{N}\Big)^L.
\label{eq:app-between-closure}
\end{equation}
Its left-hand side is fixed by the input kernel and the weight prior, so that Eq.~\eqref{eq:app-between-closure} determines $r$. For $N>C$, the right-hand side vanishes at $r=0$ and increases strictly and without bound for $r>0$, as $r$ and $(1+(r-C)/N)^L$ are positive and increasing, since $1+(r-C)/N\ge1-C/N>0$. Equation~\eqref{eq:app-between-closure} therefore has a unique positive solution $r$ for every $\mu>0$, which increases with $\mu$.

Moreover, the right-hand side equals $C$ at $r=C$ and exceeds (falls below) $r$ for $r>C$ ($r<C$), so that $r$ lies between $\mu$ and $C$ and, in particular, $r\le\max(\mu,C)$. Feature learning thus moves the terminal between-class variance $\Lambda_B=n/r$ from its NNGP value $n/\mu$ towards the target variance per output, $n/C$, amplifying it for $\mu>C$ ($t_B>1$) and suppressing it for $\mu<C$ ($t_B<1$). As $r>0$, however, $t_B-t_W=r/N>0$ for every $\mu>0$, so that NC1 decreases in every layer, at a rate that increases with $\mu$.

\subsection{Limit at fixed depth-to-width ratio}
\label{app:large-width}

We now let $N\to\infty$ at fixed $\tau=L/N$ and fixed $\mu$ and show that the compression converges to Eq.~\eqref{eq:LN}. The drive $\mu$ does not depend on $L$ if $\prod_{m=1}^Lg_m$ does not, as in Fig.~\ref{fig:linear}(a)--(c). As in Corollary~\ref{cor:scaling}, we work to leading order in $\kappa/\Lambda_W$ and $\kappa/\Lambda_B$; the corrections in Eqs.~\eqref{eq:app-tw-small-kappa} and \eqref{eq:app-tb-small-kappa} accumulate over the $L$ layers to relative order $\tau(r+C)\kappa/\Lambda_X$ and persist in this limit.

\paragraph{Rate.}
In logarithmic form, Eq.~\eqref{eq:app-between-closure} reads $\ln r+L\ln\big(1+(r-C)/N\big)=\ln\mu$. As $r\le\max(\mu,C)$ is bounded, $L\ln\big(1+(r-C)/N\big)=(r-C)\tau+O(\tau/N)$, so that for $N\to\infty$ the equation becomes
\begin{equation}
\ln r_\tau+(r_\tau-C)\,\tau=\ln\mu .
\label{eq:app-closure-tau}
\end{equation}
Its left-hand side increases from $-\infty$ to $\infty$ with slope $1/r+\tau$ in $r$, so that it has a unique root $r_\tau$, which, like $r$, lies between $\mu$ and $C$. As the slope exceeds $1/\max(\mu,C)$ there, $r=r_\tau+O(\tau/N)$. Exponentiating Eq.~\eqref{eq:app-closure-tau} and multiplying by $\tau e^{C\tau}$ gives $(r_\tau\tau)\,e^{r_\tau\tau}=\mu\tau e^{C\tau}$, an equation of the form $xe^x=z$ with $z>0$. As $xe^x\le0$ for $x\le0$ and $xe^x$ increases from $0$ to $\infty$ for $x>0$, this equation has exactly one real solution, which is positive and is denoted by $W_0(z)$, the principal branch of the Lambert function. Hence,
\begin{equation}
r_\tau=\frac{1}{\tau}\,W_0\big(\mu\tau e^{C\tau}\big).
\label{eq:app-r-tau}
\end{equation}
\paragraph{Compression.}
By Theorem~\ref{thm:geometric} and Eq.~\eqref{eq:app-transfer-ratio-expansion},
\begin{equation}
\ln\frac{\NC_1^{(L)}}{\NC_1^{(0)}}
=L\ln\frac{t_W}{t_B}
=-r_\tau\tau+O\Big(\frac{\tau}{N}\Big).
\label{eq:app-log-compression}
\end{equation}
The error term collects the corrections of order $1/N^2$ per layer in Eq.~\eqref{eq:app-transfer-ratio-expansion}, which add up to $O(\tau/N)$ over $L=\tau N$ layers as $r$ is bounded, and $(r-r_\tau)\tau=O(\tau^2/N)$. For $N\to\infty$ at fixed $\tau$ and $\mu$, it vanishes, and
Eq.~\eqref{eq:app-r-tau} gives
\begin{equation}
\lim_{N\to\infty}\frac{\NC_1^{(L)}}{\NC_1^{(0)}}
=e^{-r_\tau\tau}
=\exp\!\big[-W_0\big(\mu\tau e^{C\tau}\big)\big],
\label{eq:app-depth-width-leading}
\end{equation}
which is Eq.~\eqref{eq:LN}.

\paragraph{Small depth-to-width ratio.}
With $W_0(x)=x-x^2+O(x^3)$, Eq.~\eqref{eq:app-r-tau} gives $r_\tau=\mu+\mu(C-\mu)\tau+O(\tau^2)$. For $\mu\tau\ll1$ and $C\tau\ll1$, the rate thus reduces to $\mu$, its value for the NNGP kernels, and Eq.~\eqref{eq:app-depth-width-leading} to $1-\mu\tau$. The self-consistent change of the rate towards $C$ enters the compression only at second order in $\tau$; Eq.~\eqref{eq:app-depth-width-leading} includes it to all orders.

\paragraph{Scope.}
The limit is taken for the MAP equations, a saddle point for $N\to\infty$ at fixed $L$, whose corrections can accumulate over the layers, as the finite-width corrections of the neural tangent kernel do \citep{hanin2020finite}. For deep linear networks with one output and without observation noise, the posterior at large $P$ is determined by $LP/N$ \citep{hanin2023bayesian}, whereas $P$ does not enter Eq.~\eqref{eq:LN} at fixed $\mu$. Whether trained networks, which sample this posterior, follow Eq.~\eqref{eq:LN} beyond $LP/N\ll1$ therefore remains to be checked in future work.

\section{Input geometry of Gaussian classes}
\label{app:input-geometry}

The input kernel enters the depth law through $\NC_1^{(0)}$ and through the target drive $\mu$ of Corollary~\ref{cor:scaling}. We here compute both for Gaussian classes, $x_\alpha=m_c+\sigma\xi_\alpha$ for a sample $\alpha$ of class $c$, with fixed class centers $m_c\in\mathbb{R}^{d_0}$ and independent $\xi_\alpha\sim\mathcal N(0,I_{d_0})$, and determine when their expected input kernel is exchangeable.

\paragraph{Expected input kernel.}
As $\mathbb E[x_\alpha^\top x_\beta]=m_c^\top m_{c'}+d_0\sigma^2\delta_{\alpha\beta}$ for samples $\alpha$ of class $c$ and $\beta$ of class $c'$, the input kernel of Sec.~\ref{sec:linear} has the expectation
\begin{equation}
\mathbb E\big[K^{(0)}\big]
=\frac{g_0}{d_0}\big(YMY^\top+d_0\sigma^2I_P\big)+g_b\mathbf 1_P\mathbf 1_P^\top,
\qquad
M_{cc'}=m_c^\top m_{c'} .
\label{eq:app-expected-input-kernel}
\end{equation}
The noise and bias terms lie in $\mathcal A$, and $(YMY^\top)_{\alpha\beta}=M_{cc'}$ depends on the samples only through their classes. The expected input kernel is therefore exchangeable if and only if $M$ is invariant under all permutations $\pi$ of the classes, $M_{\pi(c)\pi(c')}=M_{cc'}$, i.e., if the centers have equal norms, $M_{cc}=a$, and equal pairwise overlaps, $M_{cc'}=b$ for $c\neq c'$. Then $M=(a-b)I_C+b\mathbf 1_C\mathbf 1_C^\top$, and with $Y\mathbf 1_C=\mathbf 1_P$ and Eq.~\eqref{eq:app-target-projector-decomposition},
\begin{equation}
YMY^\top=n(a-b)(E+P_B)+bP\,E ,
\end{equation}
whose eigenvalues are $n(a-b)$ in channel $B$ and $0$ in channel $W$. The centers $\pm e_1,\pm e_2$ of four-class XOR (App.~\ref{app:experiments}) have equal norms but overlaps $-1$ and $0$, and thus violate this condition.

\paragraph{Channel eigenvalues and SNR.}
With the signal-to-noise ratio $\SNR=\sum_c\|m_c-\bar m\|^2/(Cd_0\sigma^2)$ of Sec.~\ref{sec:linear}, where $\bar m=\frac1C\sum_cm_c$, and $\sum_c\|m_c-\bar m\|^2=\Tr(H_CM)=(C-1)(a-b)$, the channel eigenvalues of the expected input kernel are
\begin{equation}
\lambda_W^{(0)}=g_0\sigma^2,
\qquad
\lambda_B^{(0)}=g_0\sigma^2+\frac{g_0\,n(a-b)}{d_0}
=g_0\sigma^2\Big(1+\frac{nC}{C-1}\SNR\Big).
\label{eq:app-input-eigenvalues}
\end{equation}
By Eq.~\eqref{eq:app-linear-nc1-ratio}, we get
\begin{equation}
\NC_1^{(0)}=\frac{P-C}{(C-1)+nC\,\SNR}=\frac{n-1}{n\,\SNR+1-1/C}.
\label{eq:app-input-snr}
\end{equation}
For $n\to\infty$, it tends to $1/\SNR$, the NC1 of the input distribution. At finite $n$, it is smaller, as the within-class subspace has dimension $P-C$ instead of $P$ and the empirical class means retain part of the noise.

\paragraph{Target drive.}
Inserting Eq.~\eqref{eq:app-input-eigenvalues} into
Eq.~\eqref{eq:app-between-drive} gives
\begin{equation}
\mu=\frac{n}{g_0\sigma^2\big(1+\frac{nC}{C-1}\SNR\big)\prod_{m=1}^Lg_m}.
\label{eq:app-input-drive}
\end{equation}
At small $\kappa$, the rate $r$ of Corollary~\ref{cor:scaling} is an increasing function of $\mu$ and thus depends on the input only through $\lambda_B^{(0)}$, i.e., on $\sigma$ and the SNR separately rather than on the SNR alone. Raising the SNR by lowering $\sigma$ at fixed centers lowers $\lambda_B^{(0)}$ and thus raises $\mu$, $r$ and the per-layer decrease of NC1, as in Fig.~\ref{fig:linear}(b); raising it by separating the centers at fixed $\sigma$ raises $\lambda_B^{(0)}$ and has the opposite effect.

\paragraph{Sampled data.}
For general sampled inputs, $K^{(0)}$ is not exchangeable. Its orthogonal projection onto $\mathcal A$ with respect to the Frobenius inner product has the channel eigenvalues $\lambda_X^{(0)}=\Tr(P_XK^{(0)})/\operatorname{rank}(P_X)$ and thus, by Eq.~\eqref{eq:app-linear-nc1-ratio}, the same NC1 as $K^{(0)}$. The theory in Fig.~\ref{fig:linear}(d) is evaluated on this projection.

\section{Experimental and numerical details}
\label{app:experiments}

\paragraph{Synthetic data.}
We consider two balanced four-class Gaussian-mixture tasks in $\mathbb{R}^{d_0}$ with $d_0=100$, $x_\alpha=m_c+\sigma\xi_\alpha$ for the samples of class $c$ and $\xi_\alpha\sim\mathcal N(0,I_{d_0})$. For four-class XOR (Fig.~\ref{fig:nc123}), the class centers are $m_c\in\{e_1,-e_1,e_2,-e_2\}$, one class per center, whereas the Gaussian task of Fig.~\ref{fig:linear}(d) uses $m_c=e_c$, $c=1,\dots,4$. Here, $e_i$ denotes the $i$th standard unit vector of $\mathbb{R}^{d_0}$. Only the latter has centers with equal pairwise overlaps, so that its expected input kernel is exchangeable (App.~\ref{app:input-geometry}). Both datasets use $\sigma=0.4$, with $P=40$ training samples ($n=10$ per class), drawn once and shared by all trained networks.

\paragraph{Network training.}
We train the networks of Sec.~\ref{sec:linear} with one modification: the network output $h^{(L)}$ is multiplied by $\gamma^{-1/2}$. Weights and biases are initialized from zero-mean Gaussian priors, with variance $v_w/d_{\mathrm{in}}$ for a weight matrix of fan-in $d_{\mathrm{in}}$ and $v_b$ for the biases. Training uses full-batch stochastic gradient Langevin dynamics (SGLD) \citep{welling11,naveh21} on the squared-error loss $\mathcal{L}$ summed over all samples and outputs. For a parameter $\theta$, one update is
\[
\theta_{t+1} = \theta_t-\eta\big(\nabla_\theta\mathcal{L}(\theta_t)+\lambda_\theta\theta_t\big)+\sqrt{\frac{2\eta T}{\gamma}}\,\xi_t,
\qquad
\xi_t\sim\mathcal{N}(0,I).
\]
Here, $\eta=\mathrm{lr}/P$ with $\mathrm{lr}=10^{-3}$, so that the data-gradient step equals one with learning rate $\mathrm{lr}$ on the per-sample averaged loss. The weight decay is matched to the prior: $\lambda_\theta$ equals $T/\gamma$ divided by the prior variance of $\theta$, i.e., $\lambda_W=(T/\gamma)\,d_{\mathrm{in}}/v_w$ for weights and $\lambda_b=(T/\gamma)/v_b$ for biases. We train $10$ networks with independent initializations and Langevin noise on the same training samples, each for $10^5$ steps. Reported empirical values are the mean and population standard deviation over networks of the NC values at the final iterate; as the training samples are shared, layer $0$ has no spread.

The stationary distribution of these dynamics is proportional to $\exp(-\gamma\mathcal{L}/T)$ times the Gaussian prior, i.e., the posterior with Gaussian observation noise of variance $\kappa=T/(2\gamma)$. As the output multiplier $\gamma^{-1/2}$ rescales the readout, this is the posterior of Sec.~\ref{sec:linear} with $g_l=v_w$ for $l<L$, $g_L=v_w/\gamma$ and $g_b=v_b$, except that the readout bias has variance $v_b/\gamma$, so that the constant term of $K^{(L)}$ is $v_b/\gamma$ instead of $g_b$; the theory uses the same convention. Hence, $\gamma$ enters the posterior only through the readout and $\kappa$, and it sets the strength of feature learning, e.g., $\mu\propto\gamma$ in Corollary~\ref{cor:scaling}.

\paragraph{Kernel NC observables.}
At each layer, we compute the kernel $K^{(l)}$ of the trained network as defined in Sec.~\ref{sec:linear} and evaluate NC1--NC3 using the kernel expressions of Proposition~\ref{thm:kernel}. The ridge parameter of NC3 is $\kappa=T/(2\gamma)$ at every layer, matching the observation noise of the readout. We additionally report a raw-feature estimate (triangles in Fig.~\ref{fig:nc123}), obtained from the Gram matrix $ZZ^\top$ of the inputs ($l=0$) or of the post-activations $\phi(h^{(l-1)})$ ($l\geq1$), i.e., without the prefactors and the bias term of the kernels of Sec.~\ref{sec:linear}. As neither changes NC1 or NC2, the two estimates differ only in NC3, which depends on the scale of the kernel relative to the ridge. For the raw features, NC3 is evaluated with the same ridge $\kappa=T/(2\gamma)$, without rescaling.

\paragraph{Nonlinear XOR experiment (Fig.~\ref{fig:nc123}).}
For Fig.~\ref{fig:nc123}, we use four-class XOR with $C=4$, $N=256$, $L=3$, $\phi=\operatorname{erf}$, $v_w=0.5$, $v_b=0.1$, $\gamma=2$ and $T=10^{-4}$, i.e., $\kappa=2.5\times10^{-5}$. The theory curves solve the first-order MAP equations, Eqs.~\eqref{eq:app-forward-nonlinear} and \eqref{eq:app-backward-nonlinear}, jointly with the terminal condition, Eq.~\eqref{eq:app-multi-terminal}. The two-point expectations of erf and its derivatives are evaluated in closed form, and the connected four-point function by quasi-Monte Carlo, with $5\times10^4$ Sobol points for all layers and iterations. The resulting equations are solved by Anderson acceleration until the residual of $K^{(L)}$ falls below $10^{-3}$ in relative spectral norm. To improve convergence at the target finite width, the solution is annealed from the NNGP kernels at an auxiliary width $N=2\times10^4$ to $N=256$ in $50$ steps equally spaced in $N^{-1/2}$, each initialized with the solution of the previous step.

\paragraph{Linear networks with an exchangeable input kernel (Fig.~\ref{fig:linear}(a)--(c)).}
Panels~(a)--(c) use the expected input kernel, Eq.~\eqref{eq:app-expected-input-kernel}, which is exactly exchangeable, so no finite dataset is sampled and no seed averaging is required. We use $C=5$, $P=50$, $d_0=400$, $v_w=1$, $v_b=0.01$, $\gamma=2$ and $\kappa=10^{-4}$, and class centers that form a regular simplex, $\|m_c\|^2=d_0$ and $m_c^\top m_{c'}=-d_0/(C-1)$ for $c\neq c'$, so that $\SNR=1/\sigma^2$. Panels~(a) and (c) use $\sigma=1.5$, i.e., $\SNR=0.444$. Throughout these experiments, $g_l=1$ for $l<L$ and $g_L=1/2$, so that $\prod_{l=1}^Lg_l$, and hence the target drive $\mu$, does not depend on depth. Consequently, varying $(L,N)$ isolates the dependence on the ratio $\tau=L/N$.

The lines show Theorem~\ref{thm:geometric}, $\NC_1^{(l)}=\NC_1^{(0)}(t_W/t_B)^l$, with $t_W$ and $t_B$ obtained from the exact finite-$\kappa$ closure in Eq.~\eqref{eq:app-terminal-fixed-point}. For all settings shown, these scalar equations have a unique root with $t_X>0$, which we find by root finding with continuation in depth from $L=1$. As an independent check, the markers solve the full MAP forward--backward matrix equations in Eq.~\eqref{eq:exact_d-w}, by Anderson acceleration of $K^{(l)}$ and $\widetilde K^{(l)}$ until the residual of $K^{(L)}$ falls below $10^{-6}$ in relative spectral norm, with the width annealed from $N=2\times10^4$ in $20$ steps equally spaced in $N^{-1/2}$. The closed-form and full-matrix solutions agree to a relative deviation below $4\times10^{-5}$ for all widths shown, $N\in\{64,128,256,512\}$.

Panel~(a) fixes $L=16$ and compares the layerwise geometric law across widths.
Panel~(b) fixes $N=256$ and $L=16$ and varies $\SNR\in\{4,2,1,0.444,0.25,0.111\}$ through $\sigma$ at fixed centers, testing the data dependence of the contraction rate $\log(t_W/t_B)$. Panel~(c) varies depth at each width so that all curves reach $\tau=0.25$: $L\in\{1,2,4,\dots,N/4\}$ for $N\in\{64,128,256,512\}$.

\paragraph{Trained linear networks (Fig.~\ref{fig:linear}(d)).}
Panel~(d) departs from the exactly exchangeable population setting and uses the sampled Gaussian task together with SGLD-trained linear networks, with $N=256$, $L=16$, $v_w=0.5$, $v_b=0.1$, $\gamma=2$ and $T=10^{-4}$. We compare the layerwise NC1 values of the trained networks with Theorem~\ref{thm:geometric} evaluated on the channel eigenvalues $\lambda_X^{(0)}=\Tr(P_XK^{(0)})/\operatorname{rank}(P_X)$ of the sampled input kernel (see App.~\ref{app:input-geometry}), with $t_W$ and $t_B$ from Eq.~\eqref{eq:app-terminal-fixed-point}. Thus, unlike panels~(a)--(c), which compare two solutions of the same theoretical equations, panel~(d) provides a direct comparison between the analytic law and representations learned by trained finite-width networks.

\end{document}